\documentclass[12pt]{article}
\usepackage[disable]{todonotes}
\newcommand{\comment}[1]{\todo[color=green!40]{#1}}
\usepackage{layouts}

\usepackage{amsmath, nccmath}
\usepackage{amsfonts}
\usepackage{amssymb}
\usepackage{amsthm}
\usepackage{dsfont}
\newtheorem{hyp}{Hypothèse}
\newenvironment{taggedhyp}[1]
 {\renewcommand\thehyp{#1}\hyp}
 {\endhyp}
\usepackage[inline]{enumitem}
\usepackage{multirow}
\usepackage{booktabs}
\usepackage{graphicx}
\usepackage{color}
\usepackage{xcolor}

\definecolor{col1}{HTML}{0C5DA5}
\definecolor{col2}{HTML}{00B945}
\definecolor{col3}{HTML}{FF9500}
\definecolor{col4}{HTML}{FF2C00}
\definecolor{col5}{HTML}{845B97}

\newcommand{\colboxwidth}{2pt}
\newcommand{\colboxheight}{5pt}
\newcommand{\colgt}{\fcolorbox{black}{col1}{\rule{0pt}{\colboxwidth}\rule{\colboxheight}{0pt}}}
\newcommand{\colsimple}{\fcolorbox{black}{col2}{\rule{0pt}{\colboxwidth}\rule{\colboxheight}{0pt}}}
\newcommand{\colmulti}{\fcolorbox{black}{col3}{\rule{0pt}{\colboxwidth}\rule{\colboxheight}{0pt}}}
\newcommand{\colknn}{\fcolorbox{black}{col4}{\rule{0pt}{\colboxwidth}\rule{\colboxheight}{0pt}}}
\newcommand{\colneumiss}{\fcolorbox{black}{col5}{\rule{0pt}{\colboxwidth}\rule{\colboxheight}{0pt}}}

\usepackage{subcaption}
\usepackage{wrapfig}
\usepackage{placeins}

\graphicspath{{figures/}}

\usepackage[left=2.5cm,right=2.5cm,top=2.5cm,bottom=2.5cm]{geometry}
\usepackage[
    natbib=true,
    style=alphabetic,
    backend=bibtex,
    useprefix=true,
    maxcitenames=1,
    maxbibnames=3,
    citestyle=alphabetic
]{biblatex}
\AtEveryBibitem{%
  \clearfield{issn} 
  \clearfield{doi} 
  \clearfield{urldate} 
  \clearfield{urlday} 
  \clearfield{urlmonth} 
  \clearfield{urlyear} 

  \ifentrytype{online}{}{
    \clearfield{url}
  }
  \ifentrytype{preprint}{}{
    \clearfield{doi}
  }
}

\usepackage{hyperref}
\usepackage{cleveref}

\newcommand{\code}[1]{\textsf{#1}}
\newcommand{\NA}{\code{NA}}
\newcommand{\lifelines}{\href{https://github.com/CamDavidsonPilon/lifelines}{\code{lifelines}}}
\newcommand{\pycox}{\href{https://github.com/havakv/pycox}{\code{pycox}}}
\newcommand{\sklearn}{\href{https://scikit-learn.org/1.1/}{\code{scikit-learn}}}
\newcommand{\neumiss}{\href{https://github.com/marineLM/NeuMiss_sota}{\code{neumiss}}}
\newcommand{\sksurv}{\href{https://scikit-survival.readthedocs.io/en/stable/}{\code{scikit-survival}}}
\newcommand{\survivalsim}{\href{https://gitlab.inria.fr/pdufosse/survival-simulation}{\code{survivalsim}}}
\newcommand{\torchtuples}{\href{https://github.com/havakv/torchtuples}{\code{torchtuples}}}

\makeatletter
\newcommand*{\declarecommand}{%
  \@star@or@long\declare@command
}
\newcommand*{\declare@command}[1]{%
  \provide@command{#1}{}%
  \renew@command{#1}%
}
\makeatother
\newcommand{\ind}[1]{\mathds{1}\left\{#1\right\}}
\newcommand{\cond}[1]{\kappa \left(#1\right)}
\newcommand{\mathcommand}[2]{\declarecommand{#1}{\ensuremath{#2}}} 
\mathcommand{\tran}{'}
\mathcommand{\R}{\mathbb{R}}
\mathcommand{\indep}{\perp \!\!\! \perp}
\mathcommand{\evt}{\delta}  
\mathcommand{\Cind}{C}
\mathcommand{\Ctd}{C^{td}}
\mathcommand{\iBS}{iBS}
\mathcommand{\iNBLL}{iNBLL}
\mathcommand{\Proba}{\mathbb{P}}
\mathcommand{\Normal}{\mathcal{N}}
\mathcommand{\Uniform}{\mathcal{U}}
\mathcommand{\Weibull}{\mathcal{W}}
\mathcommand{\Bernoulli}{\mathcal{B}}

\begin{document}


\begin{center}
{\Large
	{\sc  Une comparaison des algorithmes d'apprentissage pour la survie avec données manquantes}
}
\bigskip

\underline{Paul Dufossé} $^{1,2}$ \& S\'ebastien Benzekry $^{1,3}$
\bigskip

{\it
$^{1}$ Équipe COMputational Pharmacology and clinical Oncology (COMPO), Inria-Inserm \\
$^{2}$ Aix-Marseille Université $^{3}$ Inria Sophia Antipolis - Méditerranée 
}
\end{center}
\bigskip


{\bf R\'esum\'e.} L'analyse de survie est un outil essentiel pour l'étude des données de santé.
Une composante inhérente à ces données est la présence de valeurs manquantes. 
Ces dernières années, de nouveaux algorithmes d'apprentissage pour la survie, basés sur les réseaux de neurones, ont été conçus.
L'objectif de ce travail est d'étudier la performance en prédiction de ces algorithmes couplés à différentes méthodes pour gérer les valeurs manquantes, sur des données simulées qui reflètent une situation rencontrée en pratique, c'est-à dire lorsque les individus peuvent être groupés selon leurs covariables.
Différents schémas de données manquantes sont étudiés.
Les résultats montrent que, sans l'ajout de variables supplémentaires, aucune méthode d'imputation n'est meilleure que les autres dans tous les cas.
La méthodologie proposée peut être utilisée pour comparer d'autres modèles de survie.
Le code en Python est accessible via le package \survivalsim. 

{\bf Mots-cl\'es.} Apprentissage machine, réseaux de neurones, données de survie, données simulées, données manquantes.

\medskip

{\bf Abstract.} Survival analysis is an essential tool for the study of health data.
An inherent component of such data is the presence of missing values.
In recent years, researchers proposed new learning algorithms for survival tasks based on neural networks.
Here, we studied the predictive performance of such algorithms coupled with different methods for handling missing values on simulated data that reflect a realistic situation, i.e., when individuals belong to unobserved clusters.
We investigated different patterns of missing data.
The results show that, without further feature engineering, no single imputation method is better than the others in all cases.
The proposed methodology can be used to compare other missing data patterns and/or survival models.
The Python code is accessible via the package \survivalsim.


\bigskip\bigskip


\section{Introduction}


Ces dernières années, de nouveaux algorithmes prédictifs pour la survie, basés sur des réseaux de neurones (RN), ont émergé.
Il a été montré que ceux-ci sont compétitifs, voire meilleurs que le modèle de Cox, pour la prédiction individuelle.
Ces nouvelles méthodes sont encore peu utilisées en pratique.
Une raison possible est qu'il est difficile de donner des garanties sur leur comportement.
Les simulations -- ou expériences numériques -- sont une manière empirique de valider ces méthodes, si l'on s'efforce de simuler des données selon des distributions réalistes~\cite{crowtherSimulatingBiologicallyPlausible2013}.

Dans le même temps, la recherche contre le cancer du poumon a vu émerger de nouveaux traitements prometteurs en immunothérapie.
Ces traitements permettent de soigner environ 20\% de patients qui sont détectés à un stade avancé (c.-à-d. métastatique).
Néanmoins, des progrès restent à faire pour identifier les patients pour lesquels l'immunothérapie est bénéfique.
C'est l'enjeu de la recherche clinique basée sur la collecte et l'analyse prédictive des biomarqueurs~\cite{ciccoliniDecipheringResponseResistance2020,benzekryMachineLearningPrediction2021}.
Les données collectées sont un cas typique de l'analyse de survie (ici~: rechute ou décès) et l'outil d'analyse plébiscité est le modèle de Cox~\cite{coxRegressionModelsLifeTables1972}, parce qu'il est facilement interprétable et vient avec des garanties théoriques.
Celui-ci reste pourtant limité pour la reconnaissance de motifs complexes, et largement dépendant du travail de modélisation préalable (\textit{feature engineering}).

Une autre caractéristique inhérente aux données de santé est la présence de données manquantes.
Une information peut être manquante pour des raisons qui dépendent de l'individu (absence ou refus de se soumettre à un examen), de la nature de l'étude (certaines données sont collectées seulement sur une période donnée, et donc pas pour tous les patients inclus dans l'étude) ou de considérations techniques (échantillons inexploitables, pannes).

L'objectif de cette étude est de présenter une méthodologie pour simuler avec des composantes aléatoires des jeux de données proches d'une situation rencontrée en pratique
puis comparer les algorithmes modernes d'apprentissage qui gèrent à la fois les données manquantes et la survie.

\paragraph{Notations}

Nous adoptons les notations suivantes~: $X$ est une variable aléatoire et $x$ sa réalisation est un vecteur colonne dans $\R^d$.
Sa transposée est $x\tran$.
La norme $\| \cdot \|$ est la norme euclidienne pour un vecteur et la norme de Frobenius pour une matrice.
En particulier, $I$ est la matrice identité.
Nous utiliserons les lois de probabilité suivantes~: 
\begin{enumerate*}[label=]
	\item $\Normal(\mu, \Sigma)$ est la loi normale multivariée de moyenne $\mu$ et matrice de variance-covariance $\Sigma$,
	\item $\Uniform$ est la loi uniforme sur $[0, 1]$,
	\item $\Bernoulli(p)$ est la loi de Bernoulli sur $\{0, 1\}$ telle que si $B \sim \Bernoulli(p)$, alors $\Proba(B = 1) = p$,
	\item et $\Weibull(\alpha)$ est la loi de Weibull de paramètre de forme $\alpha > 0$. 
\end{enumerate*}

\subsection{Cadre et modèles pour l'analyse de survie}\label{sec:sota}

Les $d$ covariables d'un individu $i$ sont modélisées par le vecteur aléatoire $X_i\in\R^d$, les temps de survie et de censure respectivement par les variables aléatoires $T_i$ et $C_i$. 
On observe $n$ individus comme la réalisation du triplet de variables aléatoires $\{X_i, Y_i, \evt_i\}$, $i=1,\dots,n$, avec $Y_i = \min(T_i, C_i)$ et $\evt_i = \ind{T_i < C_i}$.
Le modèle de Cox classique~\autocite{coxRegressionModelsLifeTables1972} fait les hypothèses suivantes sur la distribution des données~:

\begin{taggedhyp}{HPL}[Hasards proportionnels log-linéaires]\label{hyp:ph}
	Soit une variable aléatoire $X\in\R^d$ et le temps $t\in\R_{\ge 0}$, alors la fonction de hasard $h(t | X = x)$ est de la forme $h_0(t)\exp(\beta\tran x)$ avec $\beta\in\R^d$. 
\end{taggedhyp}

Cette hypothèse est plus forte que l'hypothèse des hasards proportionnels, où $\beta\tran x$ est remplacé par une fonction $f(x)$ possiblement non-linéaire.
Dans tous les cas, on suppose que les covariables $X$ impactent la survie de la même manière quel que soit le temps $t$.
Une hypothèse sur la relation entre survie et censure est souvent faite afin de prévenir certains biais d'observation~:
\todo{vérifier si les hypothèses sont OK avec ce qu'il y a dans les textbooks}
\begin{taggedhyp}{CA}[Censure aléatoire]\label{hyp:ca}
	Pour chaque individu le temps de censure est indépendant du temps de survie sachant les covariables, c'est-à-dire $T_i \indep C_i \, | X_i = x_i$.
\end{taggedhyp}

\subsection{R\'eseaux de neurones pour la survie}

Il existe un certain nombre de modèles pour la survie basés sur les réseaux de neurones (RN).
Si la plupart de ceux-ci relâchent l'hypothèse des hasards proportionnels~\cite{kvammeContinuousDiscreteTimeSurvival2019,leeDeepHitDeepLearning2018},
ce n'est toutefois pas le cas de DeepSurv~\autocite{katzmanDeepSurvPersonalizedTreatment2018} qui se restreint au modèle $h_0(t) f(x)$, apprenant simplement la fonction $\log f(x)$ à l'aide d'un RN.
Une autre différence majeure entre ces algorithmes est la modélisation des temps de survie, soit en temps continu, soit à l'aide d'une discrétisation~\cite{kvammeContinuousDiscreteTimeSurvival2019}.

\citet{kvammeTimetoEventPredictionNeural2019} fournissent, au travers de la librairie \pycox\footnote{Les modèles de RN proposés par \citet{kvammeTimetoEventPredictionNeural2019} optimisent une fonction de perte approximative qui dépend d'un hyperparamètre, le nombre d'individus inclus dans le terme de risque à un temps donné.
Les auteurs montrent de façon empirique que la performance du modèle est peu affectée par ce paramètre.} les résultats les plus complets, comparant notamment DeepSurv, DeepHit, et RSF sur des jeux de données simulées et réelles de taille variable.
La comparaison ne se fait toutefois que sur des jeux de données complets, et  comportant pour la plupart un grand nombre d'échantillons.
Notre objectif est d'étendre cette comparaison à une situation plus proche de la réalité médicale~: un nombre modéré d'individus (c.-à-d. d'échantillons) et un certain nombre de données manquantes. 

\subsection{Données manquantes}

On distingue 3 mécanismes de données manquantes dans un jeu de données réelles.~:
\begin{enumerate*}[label=]
	\item \textit{completely at random}~(MCAR),
	\item \textit{at random}~(MAR),
	\item \textit{not at random}~(MNAR).
\end{enumerate*}
En pratique, une donnée manquante correspond à une entrée d'un type spécial \NA~(pour \textit{not available}).
La prise en compte des données manquantes passe souvent par \emph{l'imputation} de celles-ci.
Il existe de nombreuses techniques d'imputations, la plupart d'entre elles supposent MCAR ou MAR.
Leur objectif est l'imputation elle-même ou \emph{l'inférence statistique}, deux paradigmes différents de la \emph{prédiction} dans le domaine de l'apprentissage statistique.

\citet{lemorvanWhatGoodImputation2021} soulignent que, pour réaliser une bonne prédiction, il n'est pas nécessaire d'imputer les données manquantes, mais seulement de prendre en compte l'information apportée par ce manque. 
Dans cette logique, \citet{lemorvanNeuMissNetworksDifferentiable2021} ont développé \emph{NeuMiss}, un module (de réseau de neurones) qui gère directement les valeurs \NA\ dans le vecteur d'entrée.
L'architecture de \emph{NeuMiss} se base sur une approximation du prédicteur de Bayes dont le coût calculatoire n'est pas prohibitif et montre de bons résultats dans divers scénarios simulés.

Enfin, si l'imputation multiple reste désirable, il demeure le choix de la technique à privilégier parmi l'arsenal disponible.
Peu d'implémentations d'algorithmes pour la survie intègrent de manière native un mécanisme pour les données manquantes.
Par exemple, si \citet{craigSurvivalStackingCasting2021} rapportent que la librairie \sksurv~\cite{polsterlScikitsurvivalLibraryTimetoEvent} gère les données manquantes, cela se fait seulement au travers du module de \sklearn~\cite{pedregosaScikitlearnMachineLearning}.
À notre connaissance, il n'existe pas d'étude rapportant les performances conjuguées d'une technique d'imputation avec un algorithme d'apprentissage pour la survie, comme cela a été fait récemment pour la classification~\cite{perez-lebelBenchmarkingMissingvaluesApproaches2022}.

\section{Méthodes de notre étude}\label{sec:study}

On décrit le modèle utilisé pour simuler les données ainsi que les paramètres choisis.

\subsection{Données simulées}

\paragraph{Covariables}

On fixe le nombre de covariables à $d=5$.
La population suit un modèle de mélange gaussien à 2 groupes, avec les paramètres statiques

\begin{equation*}
	X_{i}\,|\,G_i = k  \sim  \Normal(\mu_k, \Sigma)
	\, ,\enspace
	 k\in \{0,1\}
	\,,\;
	\Sigma = \begin{pmatrix}
		1 & c & c & 0 & 0 \\
		c & 1 & 0 & 0 & 0 \\
		c & 0 & 1 & c & 0 \\
		0 & 0 & c & 1 & c \\
		0 & 0 & 0 & c & 1 \\
	\end{pmatrix}
	\,,\;
	c = 0.5
	\enspace ,
\end{equation*}
et la variable aléatoire de groupe $G_i \sim \Bernoulli(p)$ est considérée comme non observée.
Les moyennes de chaque groupe sont tirées aléatoirement $\mu_k\in\R^d\sim\Normal(0, I)$.
La distribution retenue pour les expériences de la \cref{sec:results} est 
$\mu_1 \simeq [ 0.55,-1.46,-1.29,-1.51, 1.57]$,
$\mu_2 \simeq [ -0.98, 0.48, 0.63, 0.72, 0.91]$, ainsi $\| \mu_1 - \mu_2 \| \simeq 3.90$, et la matrice de variance-covariance a pour conditionnement $\cond{\Sigma} \simeq 13.93$.

\paragraph{Survie}

\begin{figure}
	\centering
	\includegraphics{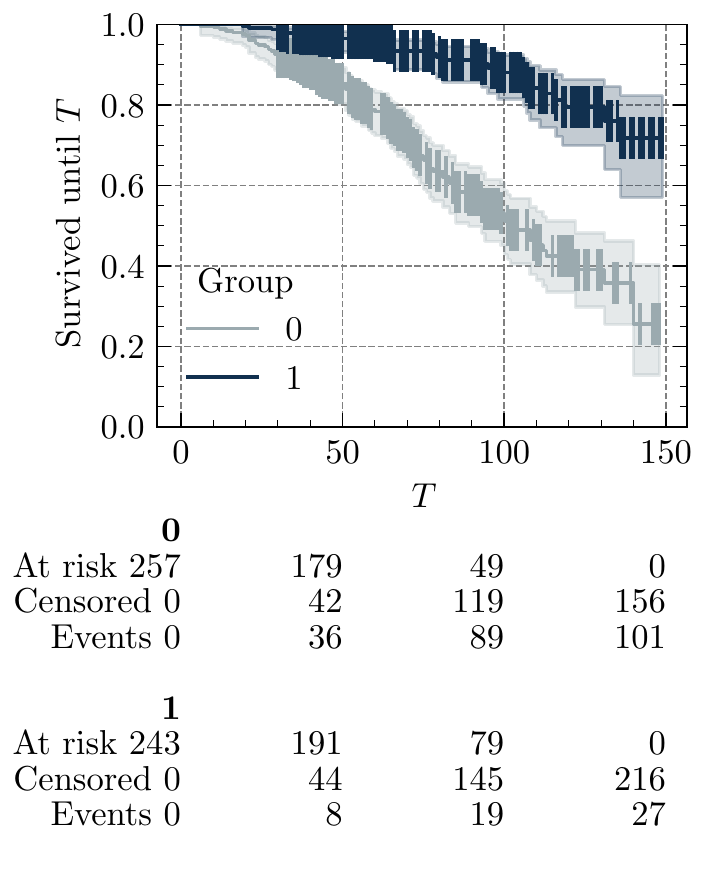}
	\caption{Courbe de Kaplan-Meier des données censurées, selon la variable $G$.}
	\label{fig:km_plot}
\end{figure}

On construit un jeu de données sans l'hypothèse \ref{hyp:ph} auquel on applique différents mécanismes de données manquantes.
Pour simuler une situation où l'hypothèse \ref{hyp:ph} n'est pas satisfaite, on propose la fonction d'interaction 
$f(x) = x_1x_2 - x_1 x_3 + 2 x_1 x_4$.
Le temps de survie $T_i$ de l'individu $i$ est généré à partir d'une loi de Weibull selon
$T_i \sim \Weibull(\alpha) \cdot \exp(f(X_i)) \cdot \lambda$.
Les paramètres de forme et d'échelle choisis sont respectivement $\alpha = 2$ et $\lambda = 100$.
Pour simuler une censure administrative, on  génère un temps de censure $C_i$ selon une loi uniforme entre 30 et 150.
Le temps de censure est donc indépendant des covariables et du temps de survie, mais la censure ($\evt_i =0)$ est plus probable pour des temps de survie plus élevés.
Ce cas est souvent rencontré en pratique lors d'une étude médicale, qui plus si elle n'est pas terminée.
Plus de détails sur la distribution des données simulées sont visibles dans la \cref{fig:km_plot} ci-contre.

\paragraph{Données manquantes}

On considère 2 mécanismes de données manquantes~: soit les données sont MCAR, selon un tirage uniforme avec une probabilité que la donnée soit manquante de $p$, soit elles sont MNAR selon un mécanisme de \emph{self-masking}~: pour chaque covariable $X^j\in\R^n$, $j=1,\dots,d$, on calcule sa moyenne empirique $\hat{X^j}$ et son écart-type $\hat{\sigma}(X^j)$, et on considère que $X_i^j$ est manquant si $|X_i^j - \hat{X^j}| > \tau \cdot \hat{\sigma}(X^j)$ avec $\tau > 0$.
Dans les résultats présentés en \cref{sec:results}, nous avons choisi $p\in\{0.4, 0.6, 0.8\}$ et $\tau\in\{0.62, 0.82, 1.03\}$ pour obtenir un taux de données manquantes d'environ (respectivement) 40, 60, et 80 \%.

\subsection{Algorithmes et métriques}\label{subsec:experimental_setup}

Nous avons choisi comme algorithmes de référence le modèle de Cox avec l'estimateur de Breslow-Aalen et les \textit{Random Survival Forests}~\autocite{ishwaranRandomSurvivalForests2008}~(RSF), auxquels s'ajoutent 2 modèles basés sur les réseaux de neurones, un continu et un discret, respectivement DeepSurv et DeepHit.
Pour gérer les données manquantes, les méthodes classiques considérées ici sont~: l'imputation simple par la médiane, l'imputation multiple au moyen d'une régression linéaire bayésienne et l'imputation par la méthode des $K$ plus proches voisins (KNN), auxquelles s'ajoute NeuMiss pour les 2 modèles de RN.
Le détail des implémentations choisies pour chaque algorithme et méthode d'imputation et leurs hyperparamètres est donné en \cref{appendix:code}.

\paragraph{Architecture et hyperparamètres des RN}
L'entrainement des réseaux de neurones impliquent de nombreux hyperparamètres~: nombre de \emph{batches, epochs}, architecture du réseau, \emph{learning rate}, \dots auxquels s'ajoutent des techniques avancées d'entrainement comme le \emph{dropout} et \emph{batch normalization}.
Nous avons utilisé la configuration par défaut du module \pycox~:
le réseau de neurones est constitué de 2 couches cachées avec 32 paramètres et la fonction d'activation non linéaire utilisée est ReLU.
Pour NeuMiss, la profondeur a été fixée à 30.

\paragraph{Choix de la métrique}
L'index \Cind\ de \citet{harrellEvaluatingYieldMedical} est une extension directe de la métrique ROC-AUC aux données censurées. C'est sans doute la métrique la plus populaire pour le problème de la survie.
Plusieurs extensions du \Cind\ ont été proposées dans la littérature biomédicale.
Par exemple, \cite{antoliniTimedependentDiscriminationIndex2005} estiment la probabilité de concordance dépendante du temps par le $\Ctd$.
Dans cette étude, nous nous sommes limités à l'inspection de $C$ parce que le modèle des données ne dépend pas du temps.
Le \Cind-index est l'estimateur de $\Proba(T_i < T_j \,\mid\, f(X_i) > f(X_j) ,\, T_i > \min(C_i, C_j))$, aussi il dépend de la censure.
Des estimateurs de la probabilité de concordance indépendante de la censure existent~\cite{gonenConcordanceProbabilityDiscriminatory2005,unoCstatisticsEvaluatingOverall2011}.
La comparaison de ces différentes métriques est en dehors du champ de la présente étude.

\paragraph{Validation croisée}
Les modèles ont été entrainés et évalués en validation croisée.
Les données sont partitionnées en 5 sous-ensembles et 4 d'entre eux forment le \emph{train} set, le dernier le \emph{test} set.
Pour l'entrainement des modèles DeepSurv et DeepHit, 20\% des données du jeu de train ont été utilisées pour former un set de \emph{validation} afin de contrôler l'\emph{overfitting}, notamment pour obtenir un critère d'arrêt (\href{https://github.com/Bjarten/early-stopping-pytorch}{\emph{early stopping}}).
De la même manière, l'imputation des données manquantes dans les jeux de test et de validation est basée seulement sur les données du jeu d'entrainement, afin d'éviter toute fuite (\emph{data leakage}).

\section{Résultats}\label{sec:results}

\begin{figure}
	\centering
		\subfloat[MCAR\label{subfig:mcar}]{\includegraphics{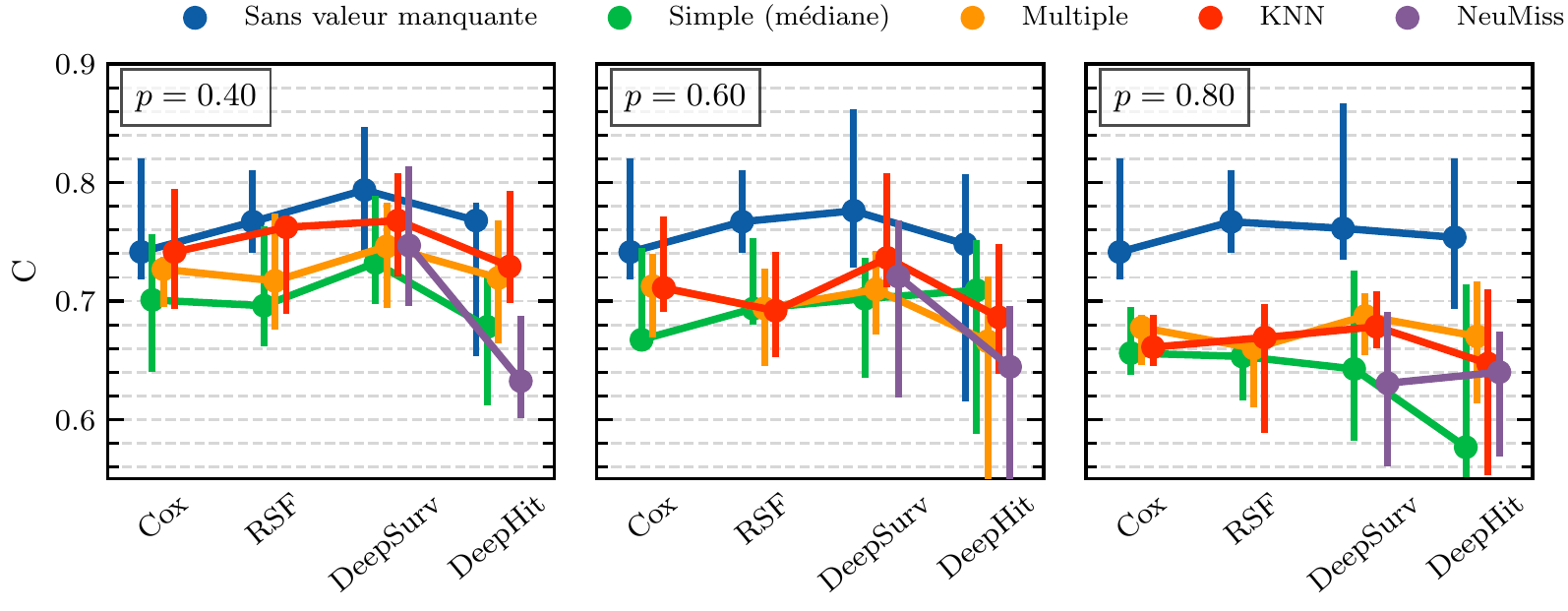}} \\
		\subfloat[MNAR\label{subfig:mnar}]{\includegraphics{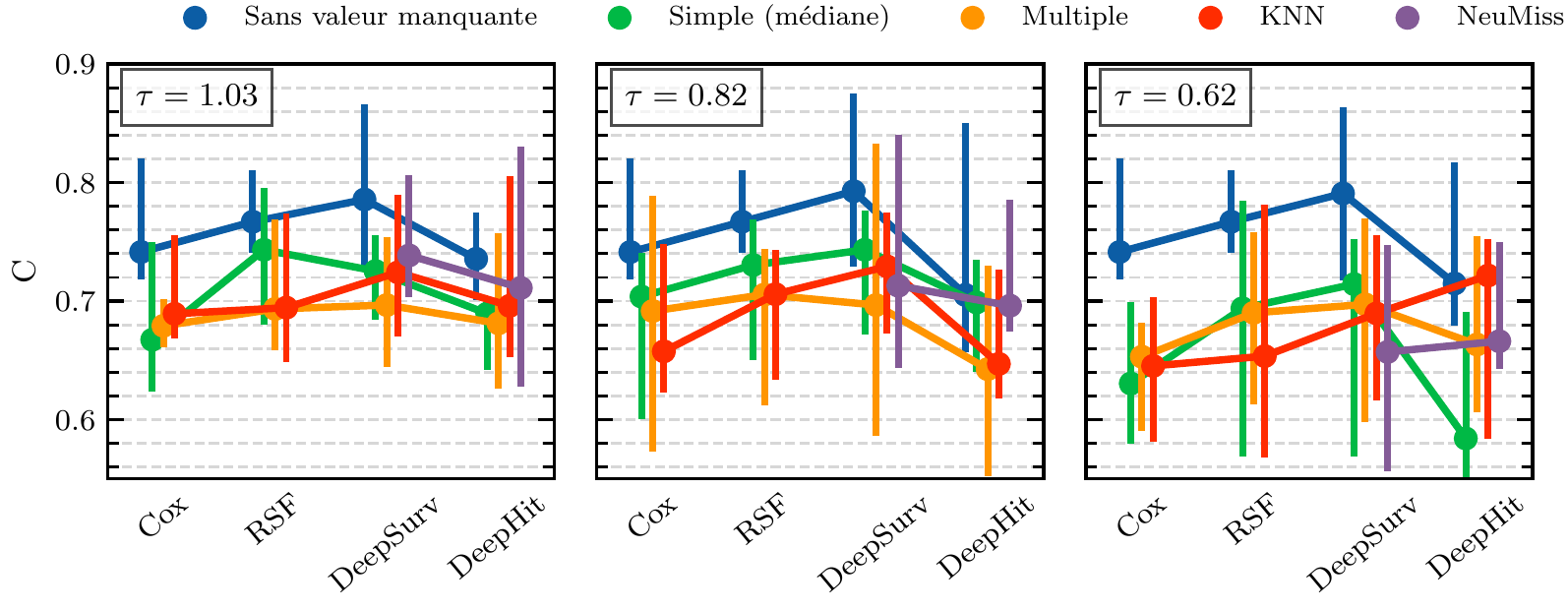}} \\
	\caption{Résultats en validation croisée avec une partition en 5 sous-ensembles, ${n=500}$ et ${d=5}$. La performance est évaluée par le $\Cind$-index (la valeur 0.5 indiquant le score d'une prédiction aléatoire).
	Le point indique la valeur médiane et les barres les valeurs extrêmes.
	Chaque jeu de données est divisé en 5 sous-ensembles.
	Les couleurs indiquent la méthode d'imputation utilisée~: sans données manquantes (\colgt), simple (\colsimple), multiple (\colmulti), KNN (\colknn) ou encore NeuMiss (\colneumiss).}
	\label{fig:results}
\end{figure}

Les expériences ont été réalisées avec $n = 500$ individus.
Les résultats sont présentés dans la \cref{fig:results}.
Une méthode est considérée comme meilleure qu'une autre lorsque la valeur médiane de la métrique est meilleure, et, dans une certaine limite, ses déviations (quantiles) ne sont pas beaucoup plus grandes.

Lorsqu'il n'y a pas de données manquantes, les performances observées suggèrent d'ordonner les algorithmes, du meilleur au moins bon~: DeepSurv, RSF, Cox et DeepHit.
Cela est attendu puisque DeepSurv est en théorie capable d'approcher le modèle non-linéaire donné par $f(x)$.
La variabilité du $C$-index en validation croisée est néanmoins plus importante pour DeepSurv que pour RSF -- le moins variable -- et le modèle de Cox, mais la valeur minimum de $C$ réalisée n'est jamais moins bonne que celle de Cox.
Nous ne sommes pas parvenus à atteindre des performances équivalentes aux autres algorithmes avec DeepHit, peut-être en raison de mauvais hyperparamètres, probablement en raison du faible nombre d'échantillons par rapport aux prérequis d'un modèle discret~\cite{kvammeTimetoEventPredictionNeural2019}.

Dans le cas MCAR~(\cref{subfig:mcar}), lorsque $p=0.4$, la meilleure méthode d'imputation est KNN et la moins bonne l'imputation simple.
Lorsque $p\in\{0.6, 0.8\}$ aucune méthode d'imputation ne se démarque, l'imputation simple restant en deçà pour le modèle de Cox.
NeuMiss est compétitif avec l'imputation multiple ou par KNN lorsque $p\in\{0.4, 0.6\}$, mais pas lorsque $p=0.8$.

Dans le cas MNAR~(\emph{self-masking}, \cref{subfig:mnar}), les combinaisons NeuMiss + DeepSurv ou RSF avec imputation simple donnent les meilleurs résultats lorsque $\tau = 1.03$ et $\tau = 0.82$ auxquelles s'ajoutent l'imputation simple et par KNN pour DeepSurv dans ce dernier cas.
Lorsque $\tau=0.62$ les résultats sont très variables, mais l'imputation simple reste performante.
Une explication est l'absence de l'utilisation du masque (colonnes additionnelles avec l'information binaire de la donnée manquante) dans les modèles étudiés.
Il a été montré que le masque améliore la performance en classification~\cite{perez-lebelBenchmarkingMissingvaluesApproaches2022}, et ici la valeur constante imputée ou NeuMiss apportent une information similaire à l'algorithme d'apprentissage.
La non-inclusion du masque et le caractère MNAR des données manquantes expliquent aussi pourquoi l'imputation multiple présente des performances médiocres.

\section{Discussion}\label{sec:conclu}

On rappelle que la comparaison des modèles n'est établie que dans la limite 
\begin{enumerate*}[label=(\roman*)]
	\item du jeu de données considéré
	\item des hyper-paramètres choisis.
\end{enumerate*}
On peut imaginer étendre ces résultats en faisant varier certains paramètres de l'étude~: le nombre d'observations, $n$, la dimension du problème (nombre de variables $d$), le modèle de données (tirages pour $G, \mu, \Sigma$ ou autre méthode de construction de clusters~\cite{qiuGenerationRandomClusters2006}), le ratio de données manquantes ou encore leur distribution.

Les résultats obtenus ici vont dans le sens de résultats récents en classification avec données manquantes~\cite{josseConsistencySupervisedLearning2019, perez-lebelBenchmarkingMissingvaluesApproaches2022}, d'autre part confirment -- dans un cas plus proche de la réalité médicale -- la supériorité de DeepSurv lorsque la fonction d'interaction avec les covariables dans la fonction de hasard n'est pas log-linéaire.
DeepSurv présente d'autres avantages applicatifs puisqu'il modélise les interactions entre covariables et permet de quantifier l'effet d'un traitement dans une population~\cite{katzmanDeepSurvPersonalizedTreatment2018}.

L'intérêt des données simulées vis-à-vis des données rencontrées en situation réelle est double~: nous disposons à la fois des données complètes -- on peut ainsi contrôler le mécanisme de données manquantes, comme il a été fait dans cette étude -- et du vrai modèle (covariables, survie, censure), ce qui peut donner lieu à l'utilisation de métriques originales pour l'entrainement. 
Dans des travaux futurs, nous aimerions exploiter cette approche pour étudier d'autres  mécanismes de données manquantes en relation avec la survie et la censure.
Par exemple, \citet{yiCoxRegressionSurvival2020} étudie le cas où l'absence des données dépend du temps de survie.
Une situation possible, lors de la collecte de données médicales, est que le praticien décide de ne pas réaliser un examen invasif, compte tenu de l'état de santé du patient.
L'état de santé est une appréciation du praticien qui peut être considérée comme une variable non observée, de laquelle dépendent à la fois le temps de survie et le mécanisme de données manquantes.

Enfin, nous prévoyons d'étendre ce genre d'étude au cas des données longitudinales.
Cela suggère notamment de prendre en compte l'approche dite \emph{survival stacking}~\autocite{craigSurvivalStackingCasting2021} qui consiste à transformer le jeu de données pour obtenir un problème de classification à partir d'observations de survie.

\FloatBarrier
\printbibliography

\appendix


\clearpage
\section{Implémentation et hyperparamètres}\label{appendix:code}

Un des objectifs secondaires de cet article est de présenter une vue sur les différentes librairies existantes en Python pour traiter le problème d'apprentissage de la survie avec données manquantes.\footnote{À cet effet, voire aussi l'annexe B de \citet{craigSurvivalStackingCasting2021} pour un inventaire des différentes méthodes d'apprentissage pour la survie, implémentées en R ou Python.}

Le modèle de Cox est implémenté dans le package \lifelines, l'algorithme RSF dans \sksurv, et les algorithmes basés sur les RN, DeepSurv et DeepHit, dans la librairie \pycox. Un code de recherche pour le module NeuMiss est accessible dans en ligne via le répertoire \neumiss.
L'imputation des données et la validation croisée sont réalisées avec \sklearn.

Les expériences réalisées dans le cadre de ce travail ont donné lieu à une librairie en Python sobrement nommée \survivalsim~(en développement) qui intègre les autres librairies mentionnées.
La \cref{table:hyperparameters} présente quelques un de ces paramètres clés, la librairie dont chacun est issu ainsi que la valeur choisie.

\begin{table}[h]
	\caption{Valeurs des hyperparamètres utilisés par les algorithmes comparés. La liste n'est pas exhaustive. Un paramètre qui n'est pas précisé conserve sa valeur par défaut.}
	\label{table:hyperparameters}
	\centering
	\begin{tabular*}{\linewidth}{@{\extracolsep{\fill}} ccccc}
		\toprule
			Module & Version & Méthode & Paramètre & Valeur \\
		\midrule
		\lifelines & 0.27.4 & \code{CoxPHFitter} & \code{penalty} & 0 puis 0.1 si \code{LinAlgError} \\
		\midrule
		\multirow{4}{*}{\sklearn} & \multirow{4}{*}{1.1} & \multirow{3}{*}{\code{IterativeImputer}} & \code{max\_iter}\footnotemark & 30 \\
		&&& \code{tol} & $10^{-2}$ \\
		&&& \code{estimator} & \code{BayesianRidge} \\
		&& \code{KNNImputer} & \code{n\_neighbors} & 10 \\
		\midrule
		\neumiss & N/A & \code{NeuMissBlock} & \code{neumiss\_depth}\footnotemark & 30\\
		\midrule
		\multirow{4}{*}{\pycox} & \multirow{4}{*}{0.2.3} & \multirow{2}{*}{\code{Model}}\footnotemark & \code{epochs} & 512 \\
		&& & \code{batch\_size} & 56 \\
		&&\multirow{2}{*}{\code{LabTransDiscreteTime}}\footnotemark & \code{cuts} & 20 \\
		&&& \code{scheme} & \code{equidistant}\footnotemark \\
		\midrule
		\multirow{4}{*}{\sksurv} & \multirow{4}{*}{0.19.0} & \multirow{3}{*}{\code{RandomSurvivalForest}} & \code{n\_estimators} & 100 \\
		&&&\code{min\_samples\_split} & 10 \\
		&&&\code{min\_samples\_leaf}  &15 \\
		&& \code{concordance\_index\_ipcw} & \code{tau} & Temps max du train \\
		\bottomrule
	\end{tabular*}
	
\end{table}
\footnotetext[3]{Donne lieu à \code{ConvergenceWarning: [...] Early stopping criterion not reached.}}
\footnotetext[4]{Valeur par défaut.}
\footnotetext[5]{\code{Model} vient en fait du package \torchtuples\ associé à \pycox.}
\footnotetext[6]{Discrétisation des temps de survies sur une grille de points équidistants}
\footnotetext[7]{La grille des temps peut être choisie de manière que les temps soient équidistants, ou selon les quantiles des temps de survie observés (méthode adaptative au problème).}

\end{document}

\section{Calcul de la probabilité de concordance}

En posant $Z_{ij} = T_i - T_j$, si $T_i \sim T_j$ (sont de même loi) et $T_i \indep T_J$ alors on peut montrer par un calcul simple que $\Proba(Z_{ij} \geq 0) = \Proba(Z_{ji} \geq 0) = 1/2$.
Le même résultat est valable pour $f_{ij} = f(X_i) - f(X_j)$ si $X_i$ et $X_j$ sont \emph{iid}.
\comment{A faire mais pas dans le papier de conf}

On calcule maintenant la probabilité suivante lorsque $X_i, X_j$ sont \textit{iid} selon la loi de l'\cref{eq:nph}~:

\begin{align}
	\Proba \left( Z_{ij} > 0 \,\mid\, f_{ij} > 0\right) &= 2 \cdot \Proba\left(Z_{ij} > 0 \,\&\, f_{ij} > 0\right) \\
	& = \int \int \ind{Z_{ij} > 0}\ind{f_{ij} > 0}d\Proba_Z d\Proba_f
\end{align}

La densité de $f_{ij}$ vaut~:

La densité de $Z_{ij}$ vaut~: